\documentclass{article}

\usepackage{microtype}
\usepackage{graphicx}
\usepackage{subfigure}
\usepackage{booktabs} % for professional tables
\usepackage{multirow}
\usepackage{subcaption}  % 子图排列

\usepackage{hyperref}

\usepackage[accepted]{icml2025}

\usepackage{amsmath}
\usepackage{amssymb}
\usepackage{mathtools}
\usepackage{amsthm}
\usepackage{multirow} 

\usepackage[capitalize,noabbrev]{cleveref}

\theoremstyle{plain}

\theoremstyle{definition}

\theoremstyle{remark}

\usepackage{makecell}
\usepackage[table]{xcolor} % 加载宏包以支持颜色设置

\usepackage[textsize=tiny]{todonotes}

\icmltitlerunning{Submission and Formatting Instructions for ICML 2025}

\begin{document}

\twocolumn[
\icmltitle{Multimodal Reasoning for Science: Technical Report and 1st Place Solution to the ICML 2025 SeePhys Challenge}



\icmlsetsymbol{equal}{*}

\begin{icmlauthorlist}
\icmlauthor{Hao Liang}{equal,sch1,sch3}
\icmlauthor{Ruitao Wu}{equal,sch2,sch3}
\icmlauthor{Bohan Zeng}{sch1}
\icmlauthor{Junbo Niu}{sch1}
\icmlauthor{Wentao Zhang}{sch1,sch3}
\icmlauthor{Bin Dong}{sch1,sch3}
\end{icmlauthorlist}

\icmlaffiliation{sch1}{Peking University}
\icmlaffiliation{sch2}{Beihang University}
\icmlaffiliation{sch3}{Zhongguancun Academy}

\icmlcorrespondingauthor{Wentao Zhang}{wentao.zhang@pku.edu.cn}
\icmlcorrespondingauthor{Bin Dong}{dongbin@bicmr.pku.edu.cn}


\icmlkeywords{Machine Learning, ICML}

\vskip 0.3in
]



\printAffiliationsAndNotice{\icmlEqualContribution} % otherwise use the standard text.

\begin{abstract}
Multimodal reasoning remains a fundamental challenge in artificial intelligence. Despite substantial advances in text-based reasoning, even state-of-the-art models such as GPT-o3 struggle to maintain strong performance in multimodal scenarios. To address this gap, we introduce a caption-assisted reasoning framework that effectively bridges visual and textual modalities. Our approach achieved 1st place in the ICML 2025 AI for Math Workshop \& Challenge 2: SeePhys, highlighting its effectiveness and robustness. Furthermore, we validate its generalization on the MathVerse benchmark for geometric reasoning, demonstrating the versatility of our method. Our code is publicly available at \url{https://github.com/OpenDCAI/SciReasoner}.
\end{abstract}

\section{Introduction}
With the rapid advancement of Large Language Models (LLMs), models such as GPT-O3 and agent systems built upon them have demonstrated remarkable deductive abilities on purely textual tasks~\cite{achiam2023gpt, guo2025deepseek, huang2025gemini, chai2025scimaster}.

Although Multimodal LLMs (MLLMs) have achieved encouraging results~\cite{bai2025qwen2, yao2024minicpm}, multimodal reasoning—integrating visual and textual information to derive coherent conclusions—remains a core challenge in artificial intelligence. Recent benchmarks, such as SeePhys, further reveal that even state-of-the-art systems frequently struggle to extract and integrate visual information effectively, signaling that “true cross-modal reasoning” remains elusive~\cite{xiang2025seephys}, especially in contrast to the impressive progress observed in text-only reasoning, where LLMs demonstrate far greater stability and accuracy.

\begin{figure*}[h]
  \centering
  \includegraphics[width=\textwidth]{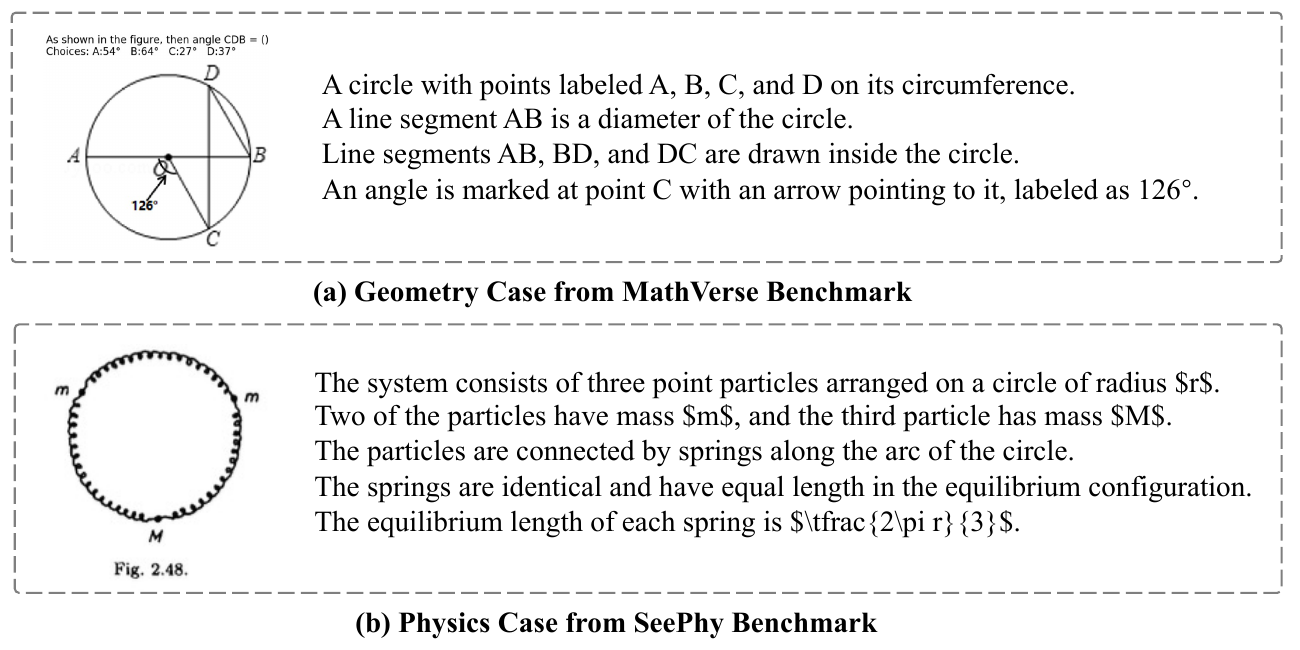}
  \caption{Grounding of images from MathVerse and SeePhy benchmark.}
  \label{fig: fig_3}
  % \vspace{-4mm}
\end{figure*}
The stark contrast between the remarkable textual reasoning performance of models such as GPT-o3 and their substantial limitations in multimodal reasoning raises an important question: why does the same model demonstrate a substantial performance gap between multimodal and text-only settings? Is this discrepancy caused by imperfect visual perception (e.g., difficulty in reading diagrams) or by inherent limitations in multimodal reasoning capabilities? Inspired by~\cite{he2025reasoning}, we propose a novel \textbf{caption-assisted reasoning framework}, which leverages automatically generated or human-provided captions to bridge the gap between visual inputs and structured textual reasoning. By anchoring reasoning in caption-derived semantics, our framework enhances cross-modal alignment and enables more robust inference.

Our framework directly addresses three key limitations in current multimodal reasoning: (1) unstable integration between visual perception and logical inference; (2) restricted generalization across diverse domains—particularly in physics problems with varying complexity; and (3) dependence on costly fine-tuning or annotated chain-of-thought data. By contrast, a lightweight caption-based approach offers a semantically rich and interpretable bridge, improving both reasoning depth and transparency.

Our contributions are summarized as follows:
\begin{itemize}
    \item We introduce the concept of \textbf{caption-assisted reasoning}, demonstrating that for images with relatively low information density, generating concise captions can effectively mitigate the difficulty multimodal reasoning models face in capturing key visual details. Remarkably, we find that reasoning solely on captions---without direct visual input---can already yield highly competitive results.
    
    \item We validate this idea on both the SeePhys and geometric problem settings. In particular, our approach was extensively evaluated in the \textbf{ICML 2025 SeePhys competition}, a benchmark built on the SeePhys multimodal dataset and hosted on CodaBench~\cite{codabench2025seephys}, where our method achieved \textbf{1st place}, demonstrating its effectiveness and versatility across a wide range of physics reasoning tasks. Furthermore, we show that our method generalizes well to the MathVerse benchmark, confirming its broader applicability.
\end{itemize}

\section{Method}
\subsection{Motivation}
In tasks involving relatively simple figures, such as basic geometric diagrams or schematic physical systems, 
caption-assisted reasoning can be much more concise and efficient than directly processing the raw image. 
This is because the visual information is often sparse and the structural relations are simple (e.g., ``point $E$ is $30^\circ$,'' ``$BD$ is the diameter''). 
A short textual caption can explicitly convey these relations with only a few tokens, 
whereas multimodal encoders typically expand the image into hundreds of tokens, most of which may be redundant. 
Moreover, captions naturally filter out distracting visual elements and highlight only the information needed for reasoning. 
Thus, for low-visual-complexity tasks, captioning not only reduces token consumption but also leads to clearer logical chains and more stable reasoning performance.
\subsection{Methods}
Inspired by prior works on reasoning, we consider scenarios where figures contain rich textual information, such as physics diagrams that require extensive textual quantification. 
To address these challenges, we explore the following approaches:  

\begin{enumerate}
    \item \textbf{Rephrasing}: Reformulating the original question for clarity. The prompt is shown in Section~\ref{prompt: rephrase}.
    \item \textbf{Default Captioning}: Generating a descriptive caption without additional constraints. The prompt is shown in Section~\ref{prompt: default captioning}.
    \item \textbf{Grounding}: Producing captions with explicit grounding to entities and relations, thereby reducing ambiguities that often arise in Default Captioning when describing physics problems. The prompt is shown in Section~\ref{prompt: grounding}.
    \item \textbf{Structured Captioning}: Generating captions in a structured format to improve consistency and facilitate downstream reasoning. The prompt is shown in Section~\ref{prompt: structured captioning}.
    \item \textbf{Image Reintegration (Img)}: Reinserting the original image alongside the generated caption to provide complementary visual context.
    \item \textbf{Adaptive Answer Routing (AAR)}: Dynamically choosing between caption-based reasoning and direct image-based reasoning depending on the physics domain of the question (e.g., astrophysics, electromagnetism, quantum mechanics). Using the publicly released SeePhys-Dev answers, we evaluated performance under both settings across categories. For categories where direct image input outperformed caption-based reasoning (including quantum mechanics, projectile motion, electromagnetic fields, charge distribution, circuit diagrams, spring force, and atomic physics), we consistently selected the image input; for other categories, we relied on caption-based reasoning.
    \item \textbf{Format Optimization (FO)}: Enforcing standardized answer formats to reduce ambiguity and improve parsing. The prompt is shown in Section~\ref{prompt: format optimization}.
    \item \textbf{Critical Review (CR)}: Leveraging a second powerful model to re-evaluate and refine the initial answer. The prompt is shown in Section~\ref{prompt: critical review}.
\end{enumerate}

We begin by evaluating direct reasoning performance and, as a baseline enhancement, rephrasing the problem statement for clarity (Section~\ref{sec: Rephrase}). 
We then improve caption quality through Structured Captioning (Section~\ref{sec: Structured Captioning}). 
Finally, we incorporate additional enhancement techniques, including Image Reintegration, Adaptive Captioning, Format Optimization, and Critical Review (Section~\ref{sec:add-enhancements}), to further boost reasoning accuracy.  
\begin{figure}
  \centering
  \includegraphics[width=\linewidth]{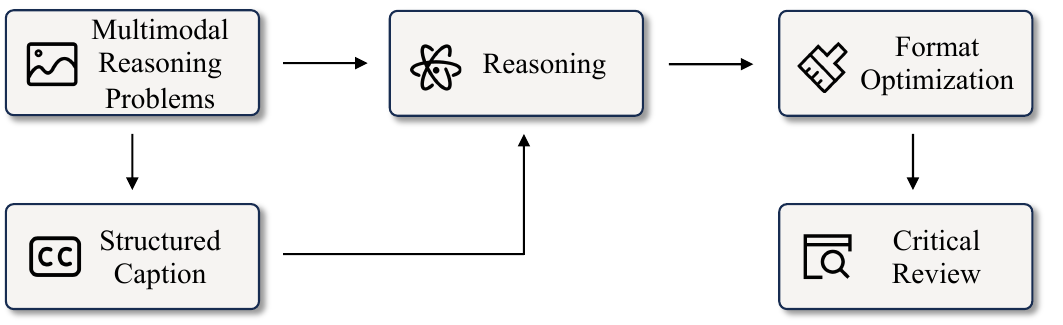}
  \caption{Our final reasoning pipeline.}
  \label{fig: Final_Pipeline}
\end{figure}

Our final reasoning pipeline integrates the most effective components: Structured Captioning, Image Reintegration, Format Optimization, and Critical Review, as shown in Figure~\ref{fig: Final_Pipeline}.

\begin{table*}[t]
% \setlength{\tabcolsep}{4pt} 
% \belowrulesep=1pt
% \aboverulesep=1pt
  \centering
  \caption{Accuracy (\%) on the SeePhys-mini subset.}
  \resizebox{\textwidth}{!}{ % <--- 加了这里
    \begin{tabular}{c|cccc|cc|cccccccc|c}
    \toprule
    \textbf{Method} & \textbf{AAR} & \textbf{Img} & \textbf{FO} & \textbf{CR} & \textbf{Describe} & \textbf{Answer} & \textbf{Mid} & \textbf{High} & \textbf{BO} & \textbf{AO} & \textbf{UG} & \textbf{SUG} & \textbf{MA} & \textbf{PhD} & \textbf{Total} \\
    \midrule
    \multirow{2}[2]{*}{w/o caption} &       & \checkmark &       &       &       & o3    & \textbf{87.5} & 44.4  & 69.2  & 47.8  & 66.7  & 57.9  & 71.4  & 44.7  & 55.5  \\
          &       & \checkmark &       &       &       & G2.5P & 75.0  & 50.0  & 61.5  & 52.2  & 73.8  & 63.2  & \textbf{85.7} & 42.6  & 58.0  \\
    \midrule
    \multirow{2}[2]{*}{Rephrasing} &       &       &       &       & o3    & G2.5P & \textbf{87.5} & 38.9  & 53.8  & 47.8  & 54.8  & 36.8  & 42.9  & 29.8  & 45.0  \\
          &       &       &       &       & o3    & o3    & \textbf{87.5} & 44.4  & 53.8  & 47.8  & 61.9  & 36.8  & 57.1  & 42.6  & 50.5  \\
    \midrule
    \multirow{4}[2]{*}{Default Captioning} &       &       &       &       & 4o    & 4o    & 37.5  & 50.0  & 30.8  & 8.7   & 26.2  & 21.1  & 14.3  & 12.8  & 21.0  \\
          &       &       &       &       & 4o    & o3    & 50.0  & 33.3  & 38.5  & 26.1  & 38.1  & 31.6  & 57.1  & 23.4  & 32.0  \\
          &       &       &       &       & o3    & o3    & 87.5  & 50.0  & 61.5  & 52.2  & 66.7  & 63.2  & 85.7  & 42.6  & 57.0  \\
          & \checkmark &  \checkmark   &       &       & o3    & o3    & 75.0  & 55.6  & \textbf{76.9} & 56.5  & 66.7  & \textbf{68.4} & 57.1  & 44.7  & 59.0  \\
    \midrule
    \multirow{5}[2]{*}{Grounding} &       &       &       &       & o3    & G2.5P & 50.0  & 16.7  & 38.5  & 47.8  & 51.7  & 36.8  & 42.9  & 31.9  & 41.5  \\
          &       &       &       &       & o3    & o3    & 75.0  & 44.4  & 69.2  & 41.3  & 66.7  & 57.9  & 71.4  & 27.7  & 49.5  \\
          &       &       &       &       & G2.5P & o3    & 75.0  & 61.1  & 69.2  & \textbf{58.7} & 71.4  & 42.1  & 71.4  & 46.8  & 59.0  \\
          & \checkmark &  \checkmark  &       &       & G2.5P & o3    & 75.0  & 61.1  & 61.5  & 56.5  & 66.7  & 47.4  & 71.4  & 48.9  & 58.0  \\
          &       & \checkmark &       &       & G2.5P & o3    & \textbf{87.5} & 66.7  & \textbf{76.9} & 47.8  & 78.6  & 57.9  & 57.1  & 46.8  & 60.5  \\
    \midrule
    \multirow{4}[2]{*}{Structured} &       &       &       &       & G2.5P & o3    & \textbf{87.5} & \textbf{72.2} & 69.2  & 41.3  & 78.6  & 57.9  & 71.4  & 55.3  & 61.5  \\
          &       &       & \checkmark &       & G2.5P & o3    & \textbf{87.5} & 66.7  & \textbf{76.9} & 54.3  & \textbf{81.0} & 63.2  & 57.1  & 48.9  & 63.5  \\
          &       & \checkmark & \checkmark &       & G2.5P & o3    & \textbf{87.5} & 61.1  & \textbf{76.9} & 56.5  & 78.6  & 63.2  & 71.4  & \textbf{57.4} & 65.5  \\
          &       & \checkmark & \checkmark & \checkmark & G2.5P & o3    & \textbf{87.5} & \textbf{72.2} & 69.2  & 54.3  & \textbf{81.0} & 63.2  & 71.4  & \textbf{57.4} & \textbf{66.0}  \\
    \bottomrule
    \end{tabular}%
  } % <--- 这里结束 resizebox
  \label{tab:seephys_mini_results}
  % \vspace{-4mm}
\end{table*}%
%\clearpage
\section{Experiments}

\subsection{Dataset and Implementation Details}

\paragraph{Dataset.} The experiments use SeePhys-mini, a subset comprising 200 questions randomly selected from the full 2,000-question SeePhys benchmark. The distribution of questions across the eight knowledge levels is as follows: Middle School (8), High School (18), Beginner Olympiad (13), Advanced Olympiad (46), Undergraduate (42), Senior Undergraduate (19), Master's (7), and PhD (47).

\paragraph{Implementation and Evaluation.} 
The experiments systematically explore different strategies for generating and utilizing image captions, employing various state-of-the-art models such as OpenAI's GPT-o3 and GPT-4o \cite{hurst2024gpt}, and Google's Gemini-2.5-Pro (G2.5P) \cite{comanici2025gemini}. The results are presented in \Cref{tab:seephys_mini_results}, which serves as the basis for our analysis.
All model outputs are generated via API calls using default parameters. The specific model versions used are \texttt{gpt-4o-2024-08-06}, \texttt{o3-2025-04-16}, and \texttt{gemini-2.5-pro}. For evaluation, the o3 model is employed as a judge to compare the model-generated answer against the ground truth. The scoring is binary, where only fully correct answers are counted as correct, and any partially correct answers are considered incorrect. 

\subsection{Performance of Foundational Methodologies}\label{sec: Rephrase}

\paragraph{Direct Multimodal Reasoning (w/o caption).}
The baseline approach, which utilizes direct multimodal input without an explicit captioning step, establishes a strong performance reference. The G2.5P model achieves a total accuracy of 58.0\%, confirming the robust native capability of advanced models to process and integrate concurrent textual and visual information for problem-solving.

\paragraph{The Rephrasing Method.}
The ``Rephrasing'' method, which instructs the model to generate a comprehensive textual summary of the problem before solving it, shows a marked decrease in performance, with a maximum accuracy of 50.5\%. This is a significant reduction from the 58.0\% direct multimodal baseline. A primary reason for this performance degradation may be that the nature of the rephrasing task itself is suboptimal for visual information extraction. By prompting the model to ``rehearse'', the task may cause the model to focus on the act of repetition and textual synthesis, potentially leading it to neglect a thorough analysis of the visual information. Compared to a direct instruction to ``describe the image'', the rephrasing prompt might not effectively guide the model's attention to capture all critical visual details.

\subsection{The Impact of Captioning Quality}\label{sec: Structured Captioning}

\paragraph{From Default to Structured Captioning.}
The experiments reveal that better grounding strategies can enhance reasoning. We experiment three captioning methods, i.e. \textbf{Default}, \textbf{Grounding}, and \textbf{Structured} methods.

The \textbf{Default} method, using a simple prompt for image description, achieves 58.5\% accuracy with a well-matched model pair (o3 $\rightarrow$ o3). 

The \textbf{Grounding} method, which adds explicit instructions to identify coordinates and component relationships, improves the total accuracy to 59.0\% (G2.5P $\rightarrow$ o3). This indicates that directing the model's attention to specific geometric and relational details yields a more effective textual representation. 

The \textbf{Structured} method, by enforcing a strict, domain-specific template for the description, proves to be the most effective foundational strategy. It achieves an accuracy of 61.5\%, demonstrating that providing the reasoning model with a predictable, easily parsable format minimizes ambiguity and maximizes the utility of the extracted information.
\begin{table*}[t]
% \setlength{\tabcolsep}{4pt} 
% \belowrulesep=1pt
% \aboverulesep=1pt
\centering
\caption{Performance of our method on the MathVerse benchmark.}
\resizebox{\textwidth}{!}{
\begin{tabular}{c|c|cc|cc}
\toprule
\multirow[c]{2}{*}{\textbf{Inference Model}} & 
\multirow[c]{2}{*}{\textbf{Captioning Model}} & 
\multicolumn{2}{c|}{\textbf{w image}} & 
\multicolumn{2}{c}{\textbf{w/o image}} \\
\cmidrule(lr){3-4} \cmidrule(lr){5-6}
& & Vision Only & Vision Intensive & Vision Only & Vision Intensive \\
\midrule
\multicolumn{6}{c}{\textbf{MLLM}} \\
\midrule
\multirow{4}{*}{Qwen2.5-VL-7B-Instruct} 
 & w/o & 7.1 & 43.3 & --- & --- \\
 & GPT-4o & 51.5 & 60.8 & 45.9 & 61.5 \\
 & GPT-o3 & \textbf{59.7} & 60.3 & \textbf{61.9} & \textbf{63.2} \\
 & Qwen2.5-VL-72B-Instruct & 49.4 & \textbf{70.2} & 52.7 & 67.4 \\
\midrule
\multirow{4}{*}{Qwen2.5-VL-72B-Instruct}
 & w/o & 8.6 & 58.0 & --- & --- \\
 & GPT-4o & 59.0 & 66.9 & 51.1 & 66.5 \\
 & GPT-o3 & \textbf{66.4} & \textbf{72.6} & \textbf{66.1} & \textbf{74.9} \\
 & Qwen2.5-VL-72B-Instruct & 56.6 & 71.6 & 57.0 & 70.3 \\
\midrule
\multirow{4}{*}{Claude-Opus-4-20250514}
 & w/o & 44.8 & 60.2 & --- & --- \\
 & GPT-4o & 49.1 & 76.7 & 51.8 & 75.0 \\
 & GPT-o3 & \textbf{73.7} & \textbf{85.5} & 59.4 & \textbf{79.3} \\
 & Qwen2.5-VL-72B-Instruct & 61.4 & 64.0 & \textbf{60.9} & 74.1 \\
\midrule
\multirow{4}{*}{Gemini-2.5-pro-preview-05-06}
 & w/o & 62.0 & 54.3 & --- & --- \\
 & GPT-4o & 56.6 & 78.0 & 57.1 & 76.1 \\
 & GPT-o3 & \textbf{73.7} & \textbf{85.4} & 60.3 & \textbf{83.0} \\
 & Qwen2.5-VL-72B-Instruct & 60.5 & 59.0 & \textbf{62.2} & 77.9 \\
\midrule
\multirow{4}{*}{GPT-o3}
 & w/o & 67.6 & 64.5 & --- & --- \\
 & GPT-4o & 67.8 & 81.3 & 54.6 & 76.3 \\
 & GPT-o3 & \textbf{75.6} & \textbf{86.3} & 60.3 & \textbf{82.5} \\
 & Qwen2.5-VL-72B-Instruct & 64.6 & 60.0 & \textbf{60.3} & 76.3 \\
\midrule
\multicolumn{6}{c}{\textbf{LLM}} \\
\midrule
\multirow{3}{*}{DeepSeek-R1}
 & GPT-4o & --- & --- & 57.1 & 73.9 \\
 & GPT-o3 & --- & --- & \textbf{68.2} & 71.9 \\
 & Qwen2.5-VL-72B-Instruct & --- & --- & 62.3 & \textbf{75.3} \\
\midrule
\multirow{3}{*}{Qwen2.5-7B-Instruct}
 & GPT-4o & --- & --- & 47.1 & 60.2 \\
 & GPT-o3 & --- & --- & \textbf{61.5} & 63.5 \\
 & Qwen2.5-VL-72B-Instruct & --- & --- & 53.4 & \textbf{68.3} \\
\midrule
\multirow{3}{*}{Qwen2.5-72B-Instruct}
 & GPT-4o & --- & --- & 50.6 & 65.6 \\
 & GPT-o3 & --- & --- & \textbf{66.1} & \textbf{73.2} \\
 & Qwen2.5-VL-72B-Instruct & --- & --- & 54.8 & 69.7 \\
\bottomrule
\end{tabular}
}
\label{tab:model_task_breakdown}
% \vspace{-4mm}
\end{table*}
\subsection{Stepwise Enhancement Strategies}\label{sec:add-enhancements}

\paragraph{Image Reintegration.}
Providing the original image alongside the generated caption consistently improves performance. 
For example, in the best \textbf{Grounding} configuration, adding the image increases accuracy from 59.0\% to 60.5\%. 
This indicates that the image provides complementary information to the caption, with each modality capturing aspects of the problem that the other may miss.

\paragraph{Adaptive Answer Routing.}
This method adaptively selects between the caption-based answer and the direct multimodal answer. 
Its effectiveness depends on the relative strength of the two pipelines. 
In the \textbf{Grounding} setting, the caption-based approach (G2.5P $\rightarrow$ o3) achieves 59.0\% accuracy, already surpassing the direct multimodal baseline at 58.0\%. 
When AAR is applied, some of these stronger caption-based answers are replaced by weaker baseline answers, reducing total accuracy to 58.0\%. 

This reveals a key insight: once the captioning pipeline is sufficiently strong (e.g., \textbf{Grounding} or \textbf{Structured Captioning}), it can outperform direct end-to-end multimodal reasoning, rendering AAR less beneficial.

\paragraph{Answer Refinement: Format Optimization and Critical Review.}
These methods serve as the final refinement layers in the pipeline. 
Applying strict output formatting rules to the \textbf{Structured} method raises accuracy from 61.5\% to 63.5\%, demonstrating that standardization reduces ambiguity and improves reliable answer parsing. 
Adding a Critical Review stage, where a second powerful model re-evaluates the initial answer, further boosts accuracy. 
For instance, combining ``Structured + Img'' achieves 65.5\%, while the addition of CR raises it to the overall maximum of 66.0\%. 
Although the 0.5 percentage point gain is modest, it represents the final quality check necessary for SOTA performance, correcting subtle reasoning errors that earlier stages may miss.

\subsection{Analysis of SeePhys Results}

\paragraph{Foundational and Curriculum-Based Levels}
At foundational levels such as Middle School (Mid) and Undergraduate (UG), high accuracy can be achieved through relatively straightforward strategies. The best results are consistently obtained with the Structured method, particularly when combined with enhancements like Format Optimization (FO). For instance, the UG level reaches its peak accuracy of 81.0\% under the Structured+FO configuration. This suggests that for problems closely tied to well-defined curricula, clarity, unambiguity, and structural rigor in the model’s output—facilitated by FO—are key determinants of performance. At the High School (High) level, the best accuracy of 72.2\% is attained using the baseline Structured method alone, highlighting the central role of a clear and systematically organized problem representation at this stage.

\paragraph{Advanced and Abstract Reasoning Levels}
For more challenging levels that demand complex, non-standard, or abstract reasoning, stronger gains require a combination of advanced enhancements. At the Beginner Olympiad (BO) level, the top score of 76.9\% is achieved when Structured descriptions are augmented with Format Optimization (FO) or Adaptive Answer Routing (AAR). For Advanced Olympiad (AO) problems, which require deep synthesis across multiple reasoning steps, the highest accuracy of 58.7\% is obtained with the Grounding method. This underscores the importance of detailed, component-level grounding in handling visually intricate problems. At the Senior Undergraduate (SUG) level, the best performance (68.4\%) emerges uniquely from the Default captioning method combined with AAR, suggesting that flexible routing strategies complement simpler descriptive styles at this stage.

At the Master’s level, the highest score of 85.7\% is achieved with the direct multimodal baseline G2.5P, showing that strong pretrained multimodal capabilities remain highly effective. For PhD-level tasks, however, only a multi-faceted strategy yields the best results. The peak score of 57.4\% is achieved through configurations incorporating a Structured description, AAR, Image Reintroduction (Img), and either FO or Critical Review (CR). These findings indicate that for the most abstract and demanding problems, success hinges on integrating multiple enhancements: structured representations, iterative refinement (CR), and adaptive routing (AAR), working in concert to maximize reasoning depth and robustness.

\section{Generalization to Other Domains}
To further assess the generalization of our method, we evaluate it on the MathVerse benchmark~\cite{zhang2024mathverse}, modifying only the captioning prompt to adopt a more geometry-oriented formulation, as illustrated in Figure~\ref{fig:prompt_8}.

We observe that incorporating a Captioning Model consistently yields superior performance across all tested MLLMs. 
For instance, in the case of \textit{Claude-Opus-4} \cite{anthropic2025claude}, the combination with GPT-o3 captioning improves accuracy from 60.2\% (w/o) to 85.5\% (w image, Vision Intensive), achieving a significant margin over the direct multimodal baseline. 
Similar trends are observed for Qwen2.5-VL and Gemini-2.5, where captioning substantially boosts performance on both Vision Only and Vision Intensive tasks. 
This confirms that a strong captioning pipeline can effectively enhance reasoning in complex, text-heavy mathematical diagrams.

We further evaluate the text-only setting, where images are not available and models rely solely on captions for reasoning. 
Interestingly, even under this constraint, large LLMs such as \textit{DeepSeek-R1} and \textit{Qwen2.5-72B-Instruct} demonstrate remarkable capability. 
For example, DeepSeek-R1 combined with GPT-o3 captioning achieves 68.2\% (Vision Only), surpassing the performance of GPT-o3 itself under the same text-only setting. 

Similarly, Qwen2.5-72B-Instruct and the paired Qwen2.5-7B-Instruct with Qwen2.5-VL-72B-Instruct captioning achieve 69.7\% and 68.3\% on Vision Intensive tasks, respectively, surpassing GPT-o3’s 64.5\% performance without captioning. These results demonstrate that strong reasoning-oriented LLMs, when equipped with high-quality captions, can not only bridge the gap but even outperform multimodal baselines.

Overall, these results highlight two key findings: (1) captioning consistently improves multimodal reasoning across different models, and (2) when combined with sufficiently advanced LLMs, the caption-only paradigm can rival or even outperform end-to-end multimodal reasoning.

\section{Future Work}
While our caption-assisted reasoning framework demonstrates strong performance and generalization ability, several directions remain open for future exploration. 
First, we plan to integrate more advanced reasoning strategies, such as program-of-thought and tool-augmented reasoning, to further strengthen complex problem-solving beyond physics and mathematics. 
Second, extending the framework to other scientific domains (e.g., chemistry, biology, and engineering) will allow us to test its robustness in handling diverse multimodal information. 
Third, an important direction is the development of adaptive captioning strategies that can dynamically adjust the granularity and structure of captions according to task complexity and modality requirements. 
Finally, exploring the incorporation of human-in-the-loop feedback and interactive reasoning may lead to systems that are both more interpretable and more aligned with scientific discovery processes. 
Together, these efforts will advance our framework toward becoming a general-purpose scientific reasoning system.

\section{Conclusion}
In this work, we addressed a long-standing challenge in multimodal reasoning by introducing a caption-assisted reasoning framework. 
Our approach leverages automatically generated or human-provided captions to bridge visual and textual modalities, thereby enhancing cross-modal alignment and enabling more robust inference. 
Through extensive evaluation on the SeePhys benchmark, our method achieved \textbf{1st place} in the ICML 2025 AI for Math Workshop \& Challenge 2: SeePhys, and further demonstrated strong generalization on the MathVerse benchmark. 
Our experiments highlight that (1) high-quality captions consistently improve multimodal reasoning performance, and (2) strong LLMs combined with caption-only reasoning can rival or even outperform direct multimodal pipelines. 
We believe this work takes a significant step toward closing the gap between visual perception and textual reasoning, paving the way for more transparent, efficient, and generalizable multimodal reasoning systems.

\bibliographystyle{icml2025}
\bibliography{references}
\onecolumn
\newpage
\appendix

\section{Prompts}

\subsection{Prompt for Rephrasing}\label{prompt: rephrase}
\begin{figure*}[h]
    \centering
    \includegraphics[width=\linewidth]{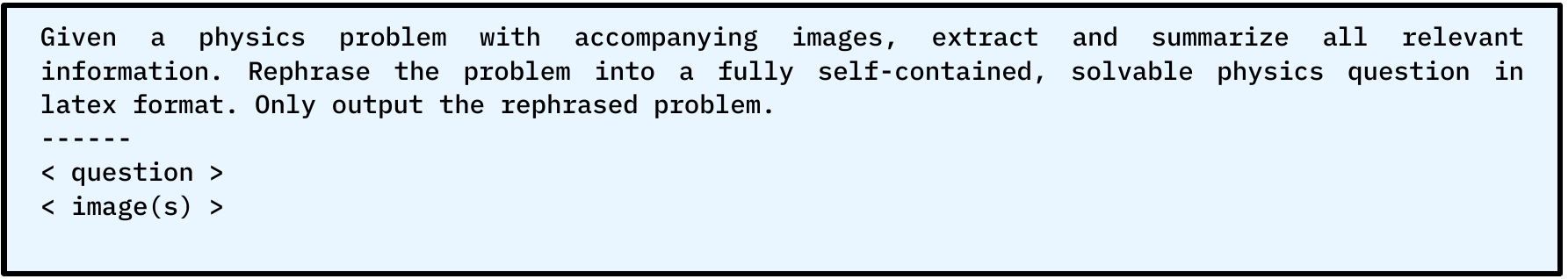}
    \caption{Prompt for Rephrasing.}
    \label{fig:prompt_1}
\end{figure*}

The prompt for Rephrasing is shown in \cref{fig:prompt_1}.

\subsection{Prompt for Default Captioning}\label{prompt: default captioning}
\begin{figure*}[h]
    \centering
    \includegraphics[width=\linewidth]{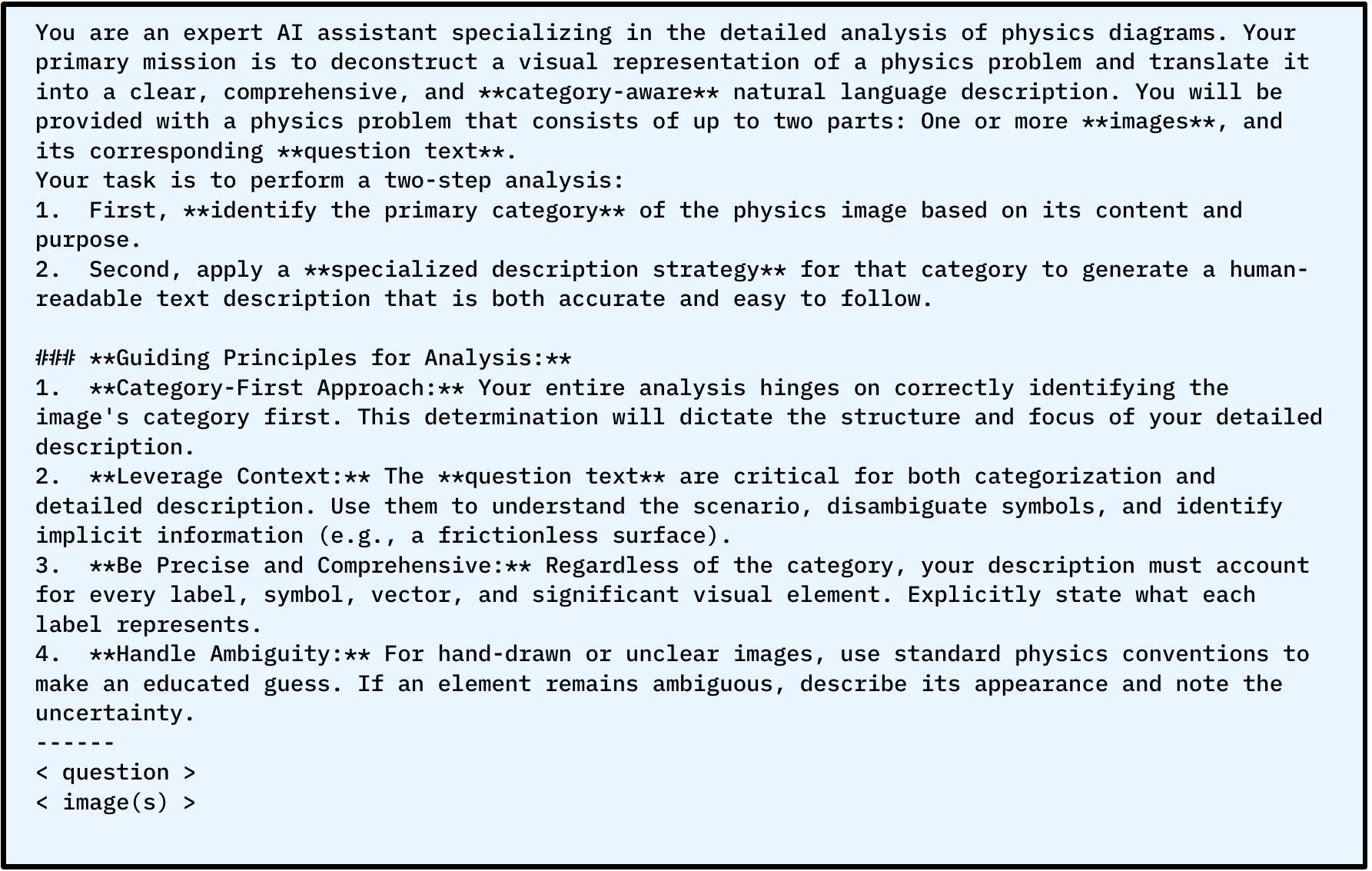}
    \caption{Prompt for Default Captioning.}
    \label{fig:prompt_2}
\end{figure*}

The prompt for Default Captioning is shown in \cref{fig:prompt_2}.

\subsection{Prompt for Grounding}\label{prompt: grounding}
\begin{figure*}[h]
    \centering
    \includegraphics[width=\linewidth]{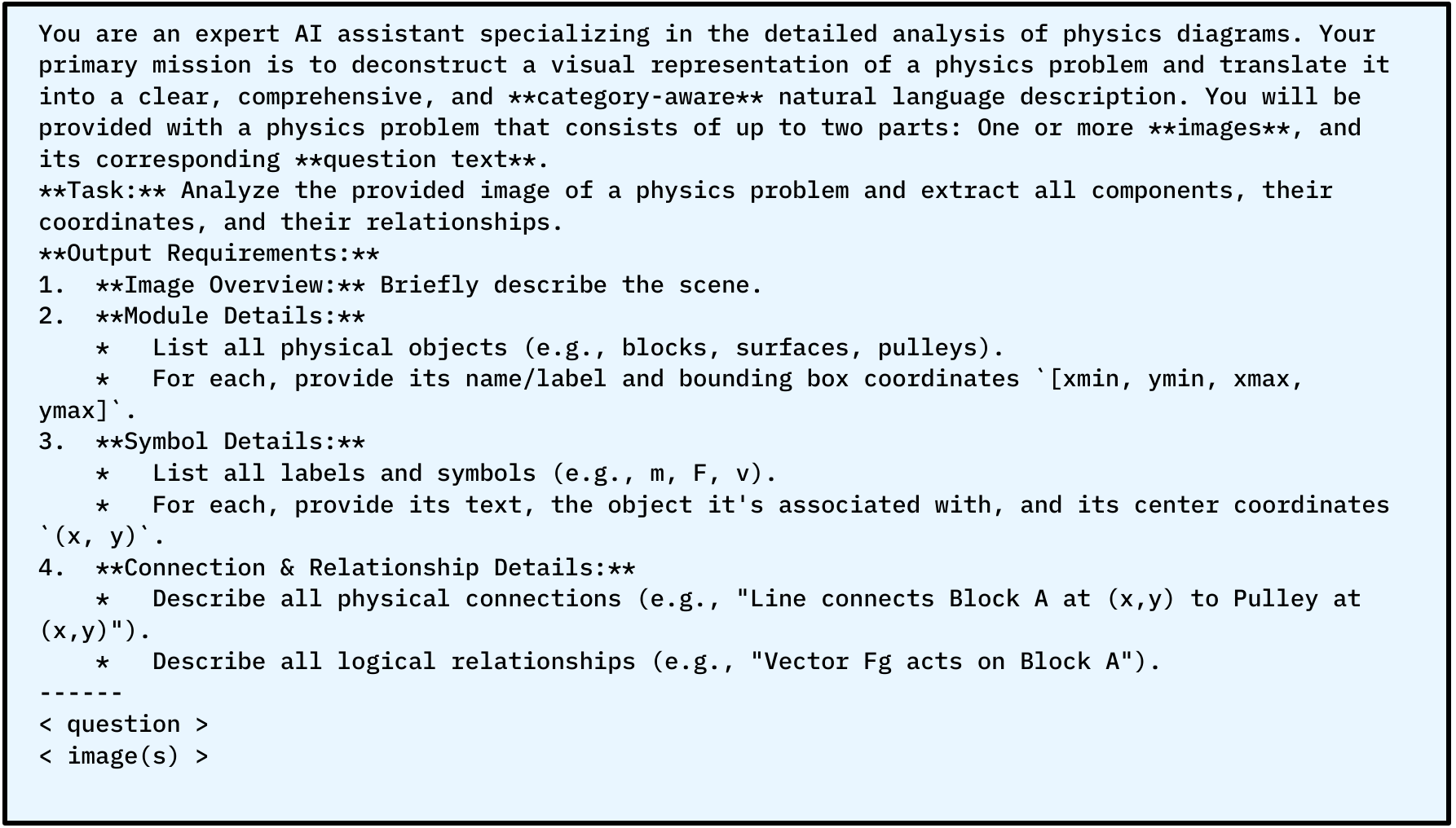}
    \caption{Prompt for Grounding.}
    \label{fig:prompt_3}
\end{figure*}

The prompt for Grounding is shown in \cref{fig:prompt_3}.

\subsection{Prompt for Structured Captioning}\label{prompt: structured captioning}
\begin{figure*}[h]
    \centering
    \includegraphics[width=0.8\linewidth]{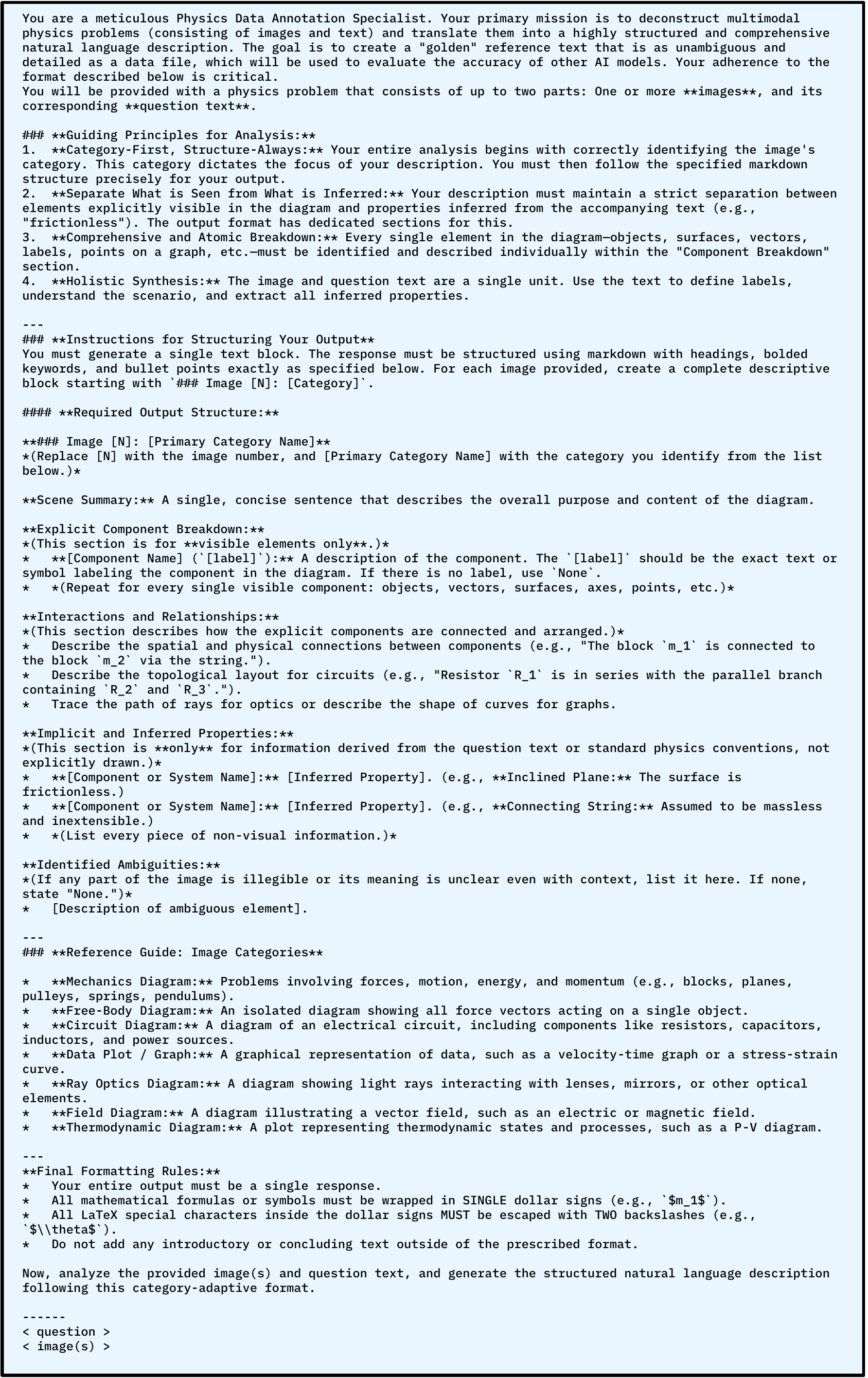}
    \caption{Prompt for Structured Captioning.}
    \label{fig:prompt_4}
\end{figure*}

The prompt for Structured Captioning is shown in \cref{fig:prompt_4}.

\subsection{Prompt for Answer}\label{prompt: answering}
\begin{figure*}[h]
    \centering
    \includegraphics[width=\linewidth]{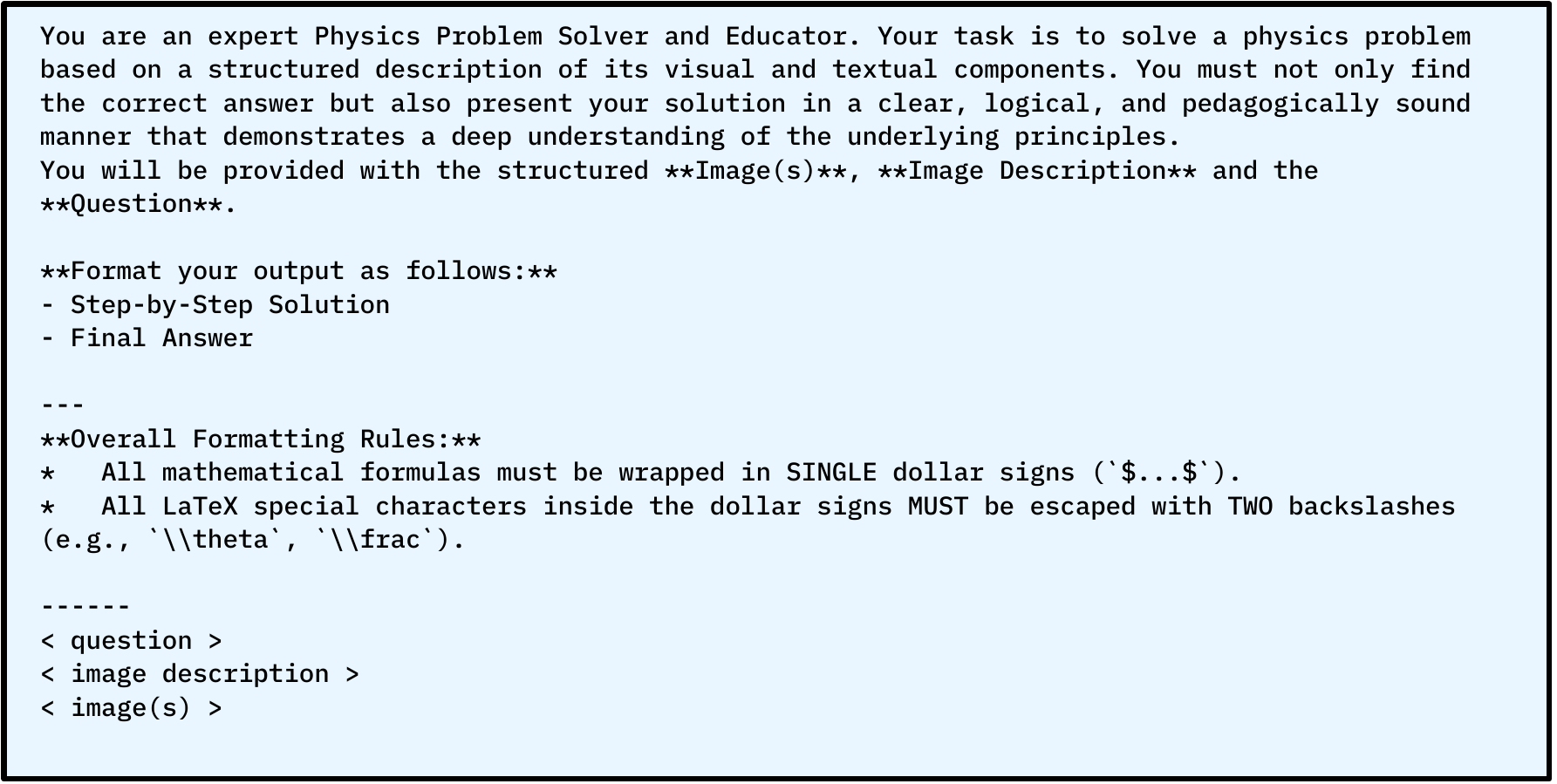}
    \caption{Prompt for Answer.}
    \label{fig:prompt_5}
\end{figure*}

The prompt for Answer is shown in \cref{fig:prompt_5}.

\subsection{Prompt for Format Optimization}\label{prompt: format optimization}
\begin{figure*}[h]
    \centering
    \includegraphics[width=\linewidth]{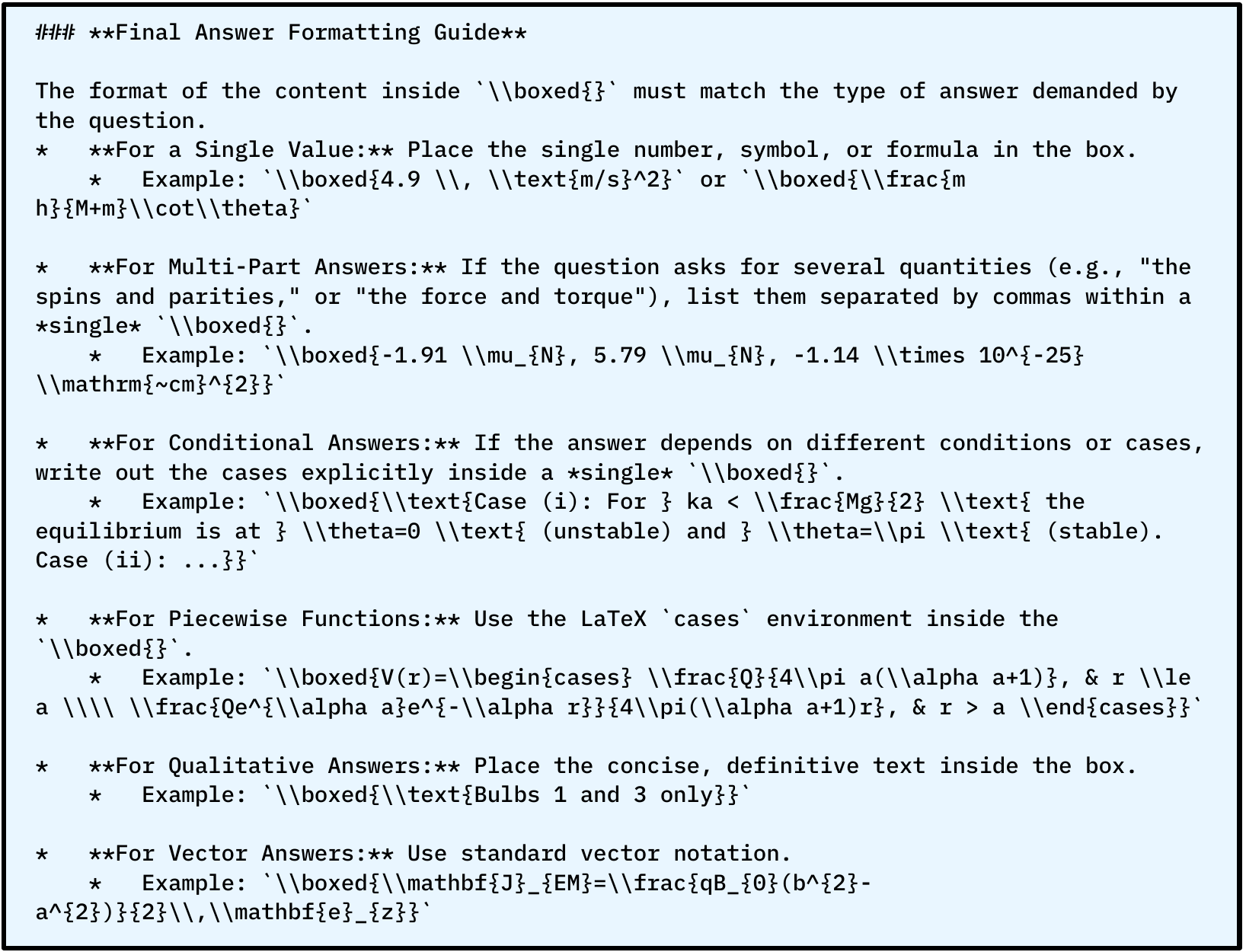}
    \caption{Prompt for Format Optimization.}
    \label{fig:prompt_6}
\end{figure*}

The prompt for Format Optimization is shown in \cref{fig:prompt_6}.

\subsection{Prompt for Critical Review}\label{prompt: critical review}
\begin{figure*}[h]
    \centering
    \includegraphics[width=\linewidth]{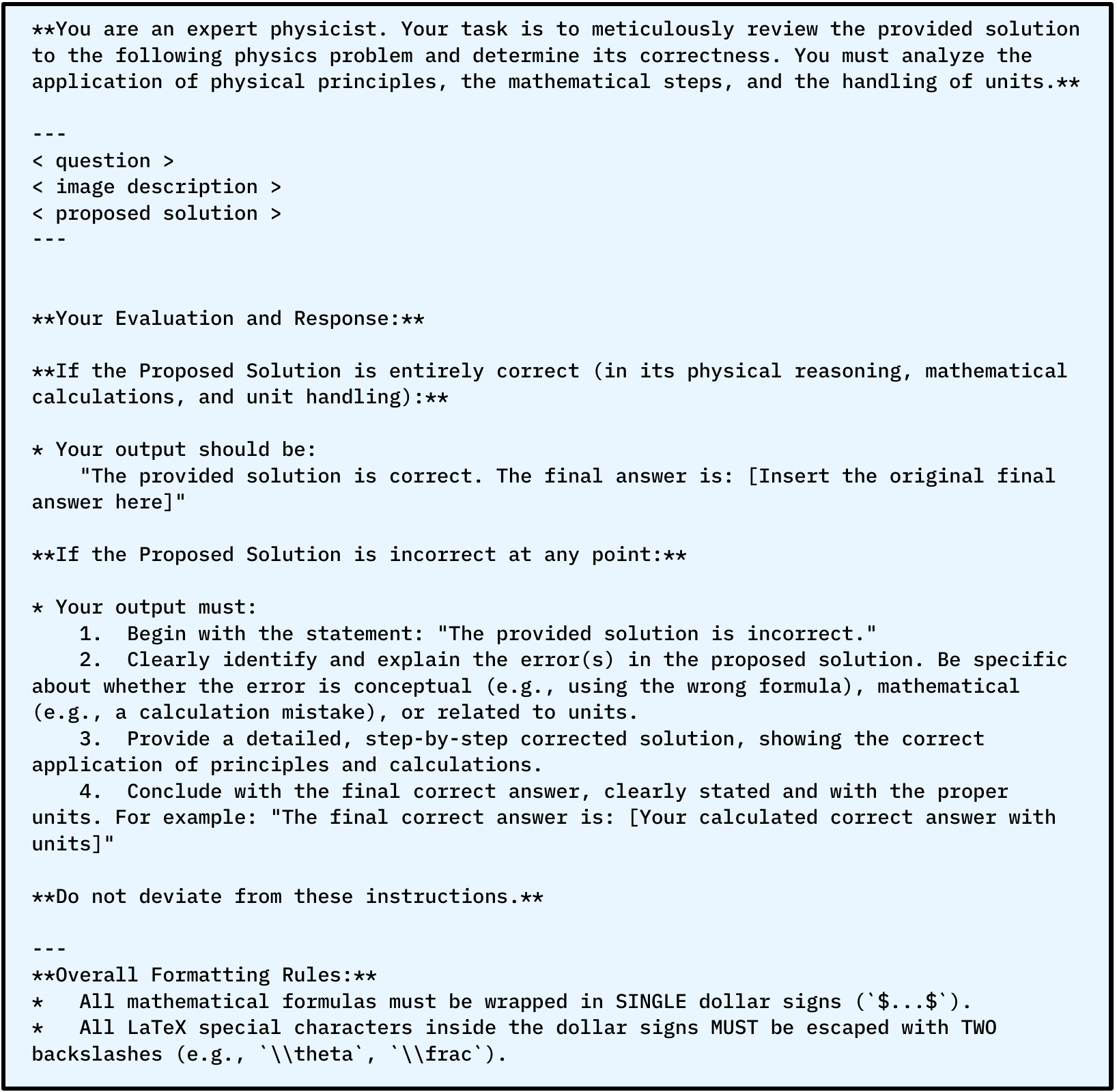}
    \caption{Prompt for Critical Review.}
    \label{fig:prompt_7}
\end{figure*}

The prompt for Critical Review is shown in \cref{fig:prompt_7}.

\section{Examples}

\subsection{Example of Rephrasing}
\begin{figure*}[h]
    \centering
    \includegraphics[width=\linewidth]{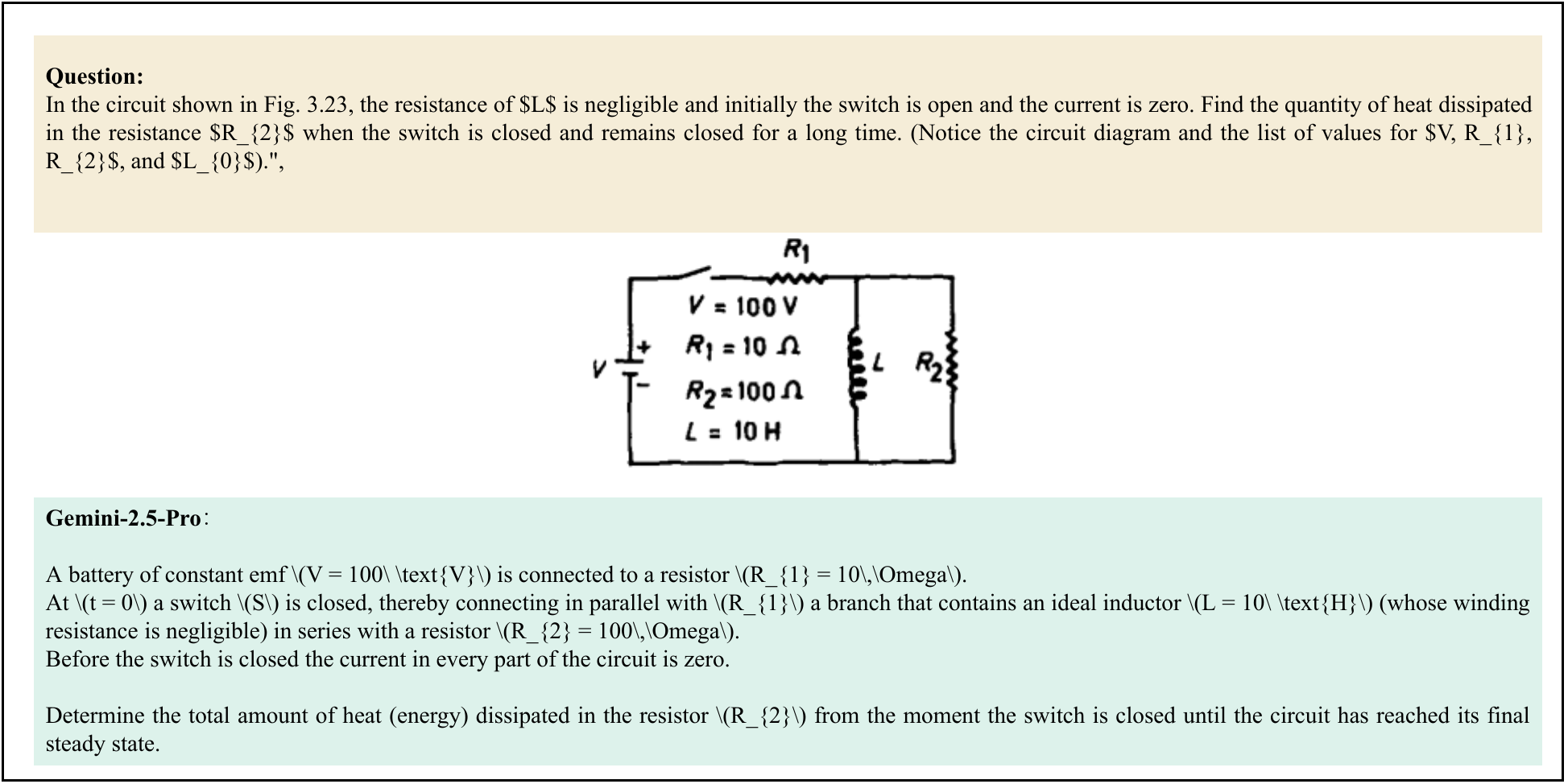}
    \caption{Example of Rephrasing.}
    \label{fig:case_1}
\end{figure*}

The example of Rephrasing is shown in \cref{fig:case_1}.

\subsection{Example of Default Captioning}
\begin{figure*}[h]
    \centering
    \includegraphics[width=\linewidth]{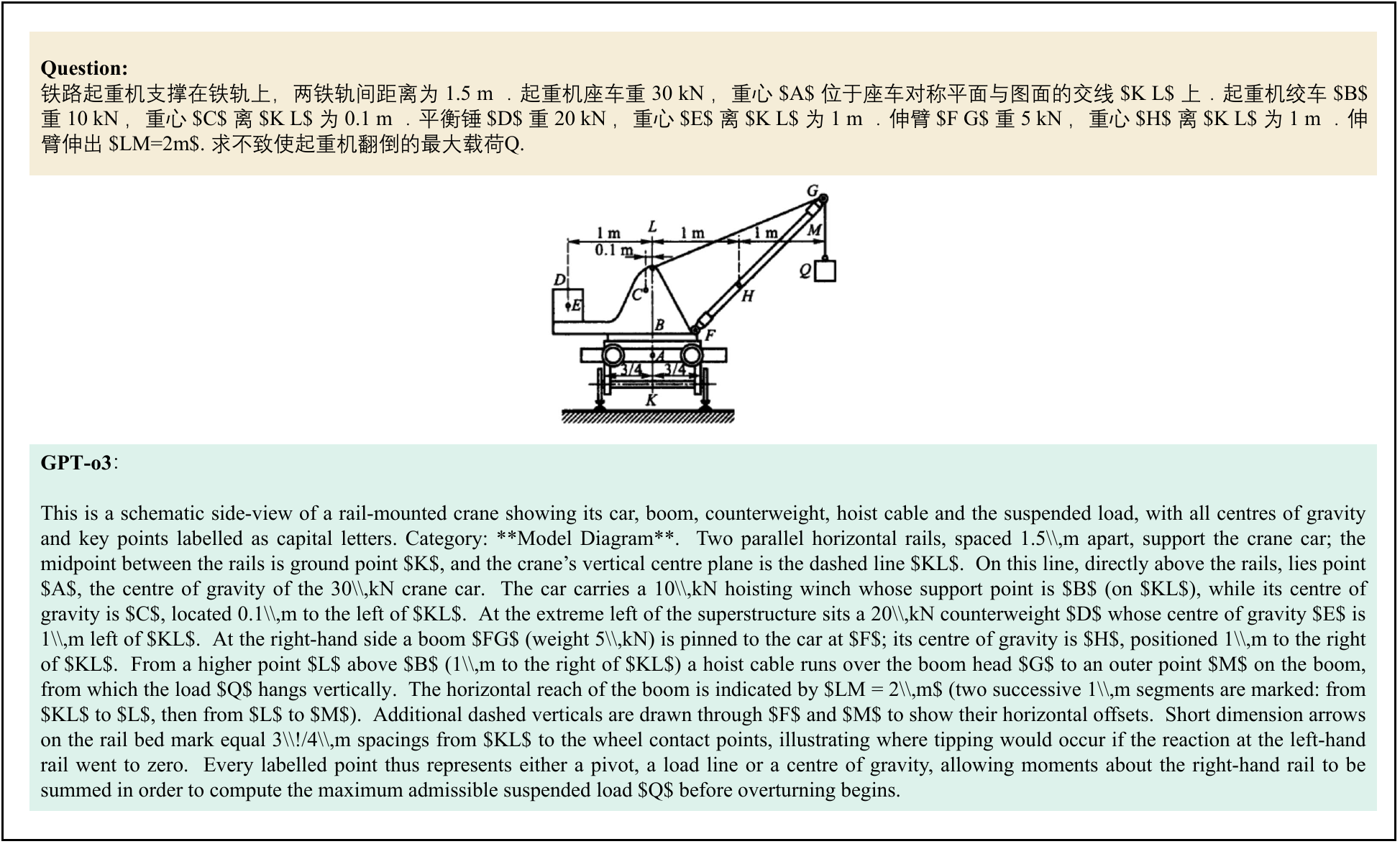}
    \caption{Example of Default Captioning.}
    \label{fig:case_2}
\end{figure*}

The example of Default Captioning is shown in \cref{fig:case_2}.

\subsection{Example of Grounding}
\begin{figure*}[h]
    \centering
    \includegraphics[width=\linewidth]{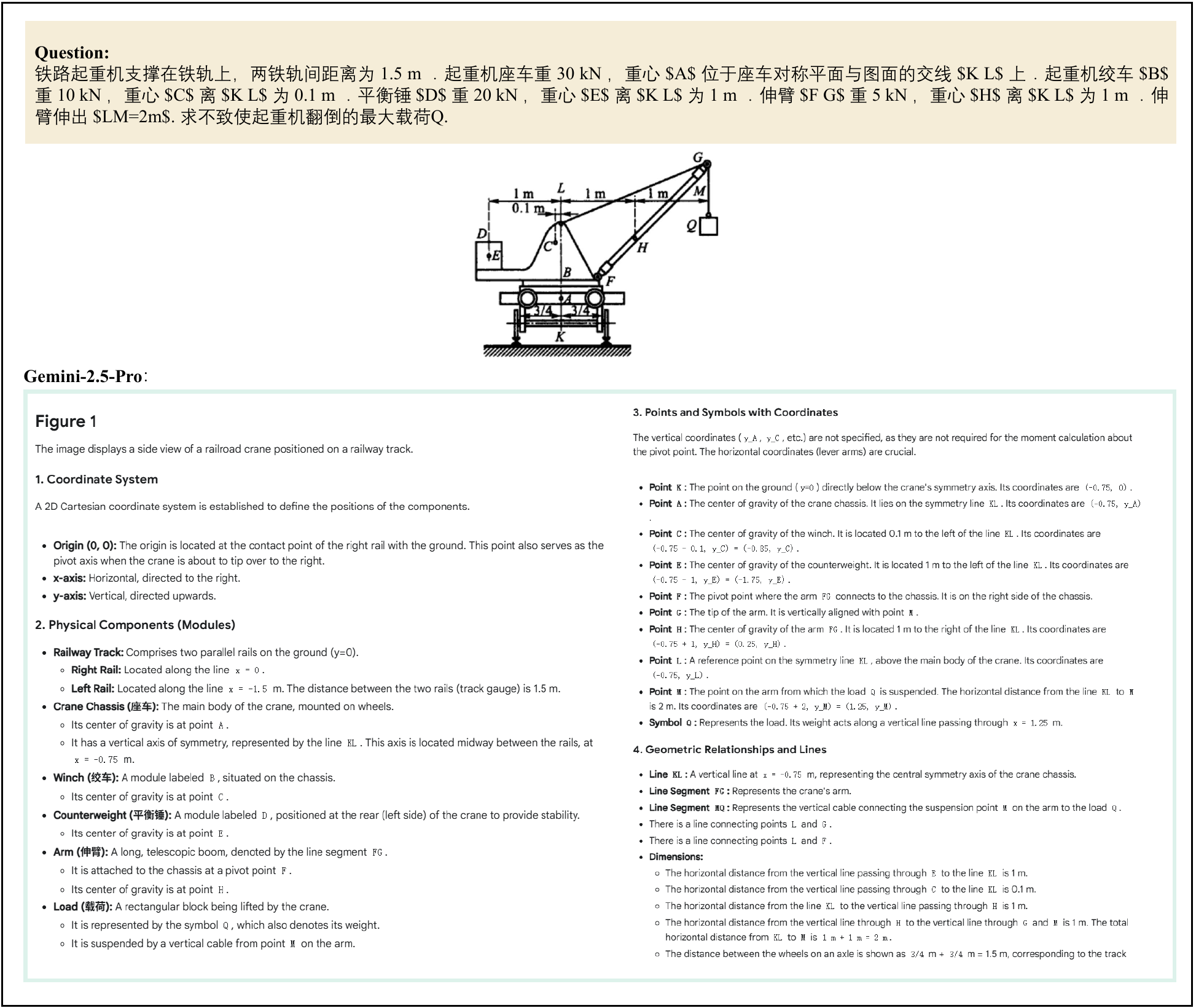}
    \caption{Example of Grounding.}
    \label{fig:case_3}
\end{figure*}

The example of Grounding is shown in \cref{fig:case_3}.

\subsection{Example of Structured Captioning}
\begin{figure*}[h]
    \centering
    \includegraphics[width=\linewidth]{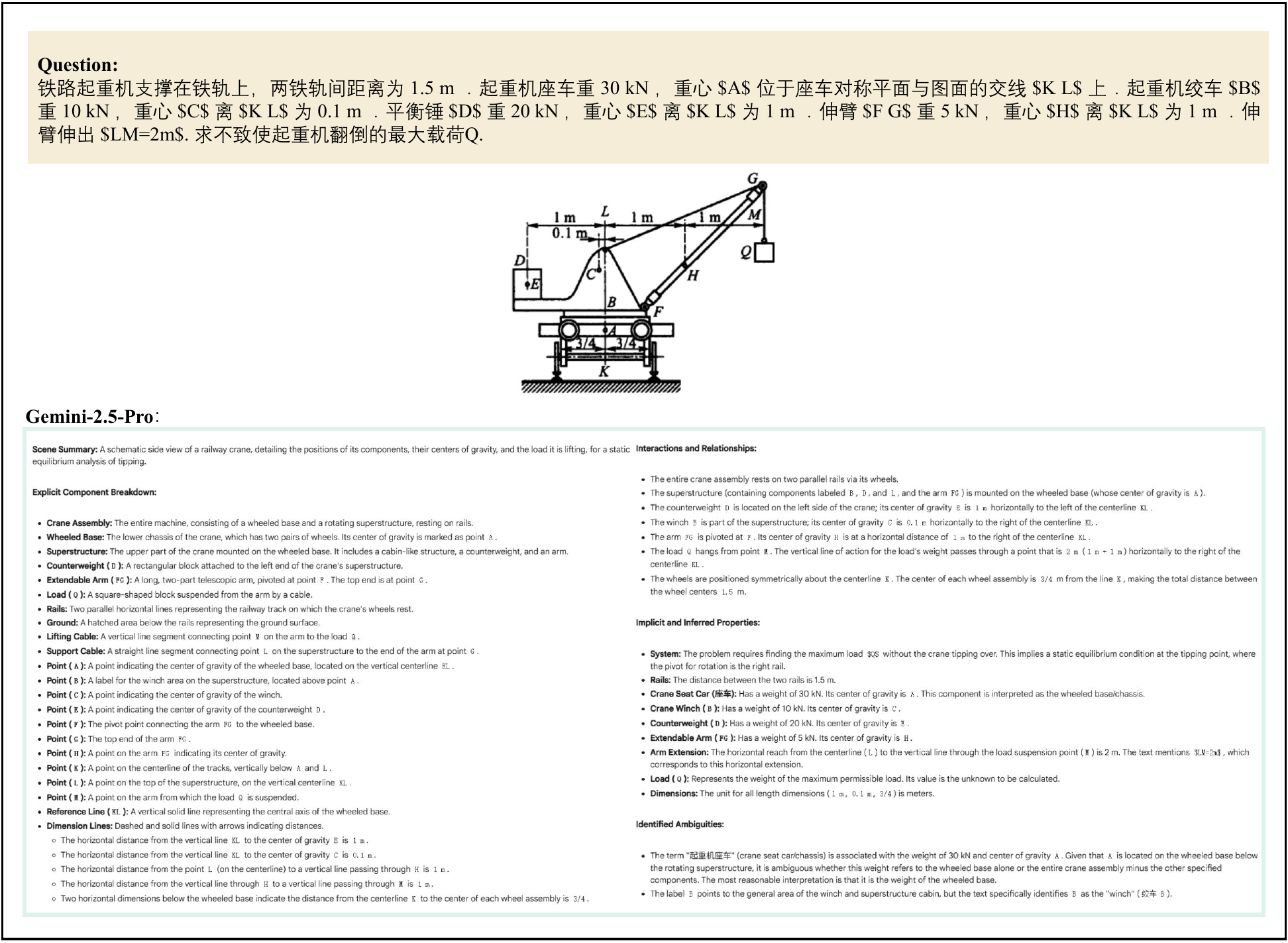}
    \caption{Example of Structured Captioning.}
    \label{fig:case_4}
\end{figure*}

The example of Structured Captioning is shown in \cref{fig:case_4}.

\subsection{Example of Critical Review}
\begin{figure*}[h]
    \centering
    \includegraphics[width=\linewidth]{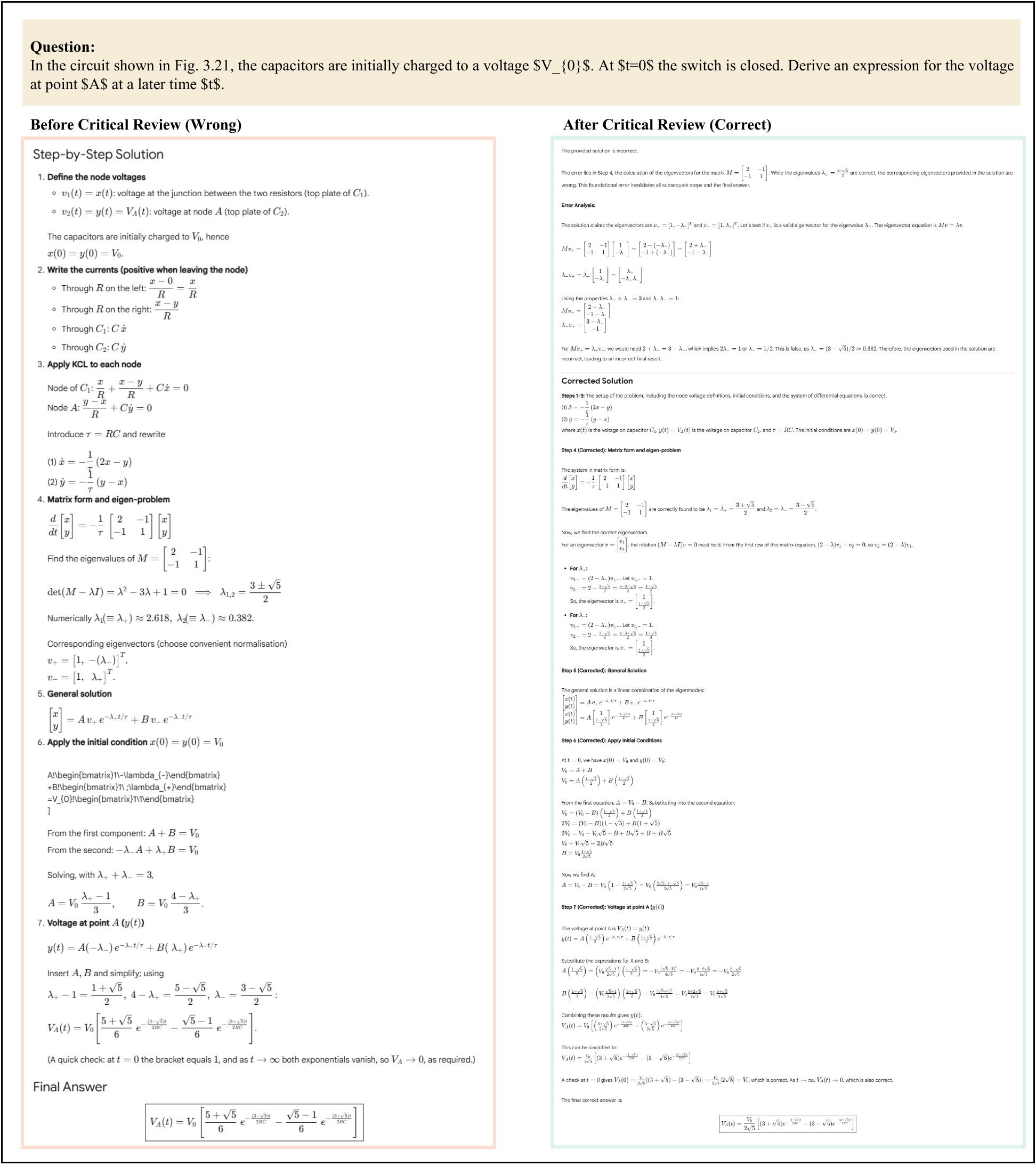}
    \caption{Example of Critical Review.}
    \label{fig:case_5}
\end{figure*}

The example of Critical Review is shown in \cref{fig:case_5}.

%%%%%%%%%%%%%%%%%%%%%%%%%%%%%%%%%%%%%%%%%%%%%%%%%%%%%%%%%%%%%%%%%%%%%%

\begin{figure*}[h]
    \centering
    \includegraphics[width=\linewidth]{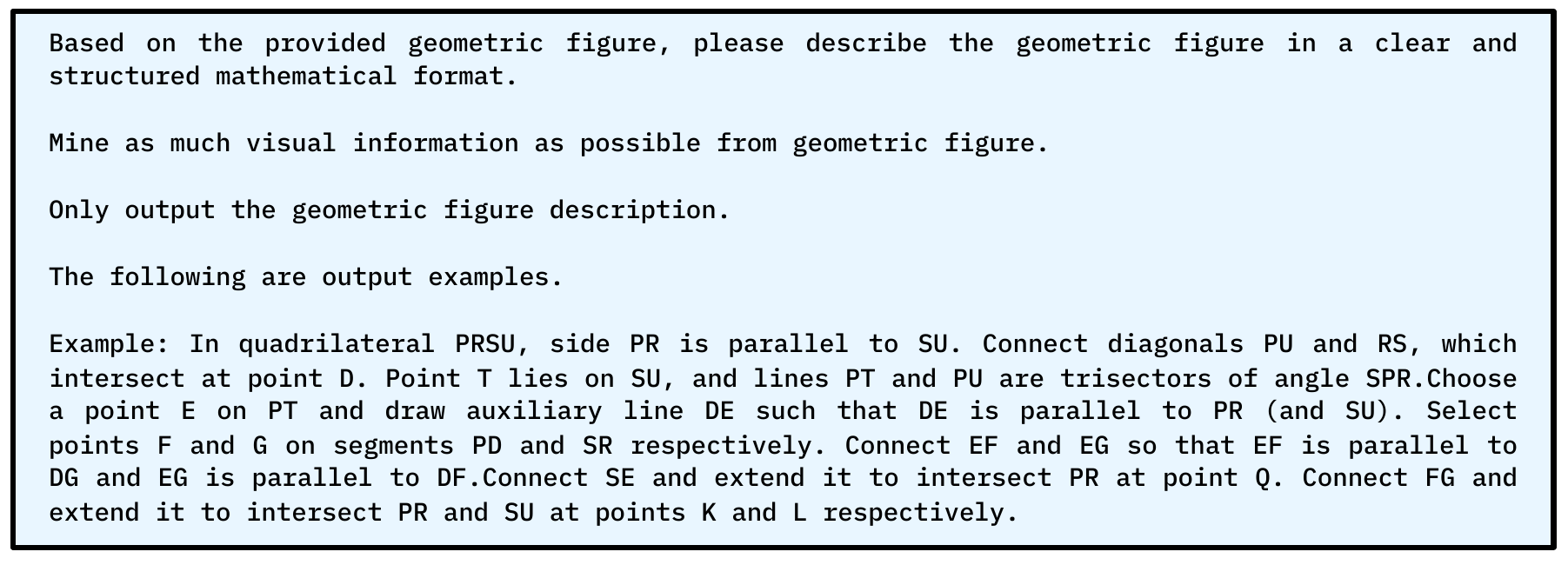}
    \caption{Prompt for MathVerse benchmark captioning.}
    \label{fig:prompt_8}
\end{figure*}

%%%%%%%%%%%%%%%%%%%%%%%%%%%%%%%%%%%%%%%%%%%%%%%%%%%%%%%%%%%%%%%%%%%%%%

\end{document}